%% file: acl_latex.tex
\documentclass[11pt]{article}

\usepackage[final]{acl}

\usepackage{times}
\usepackage{latexsym}
\usepackage{amsmath}
\usepackage{amssymb}
\usepackage{bbm}
\usepackage{makecell}
\usepackage{multirow}
\usepackage{kotex}
\usepackage{array}
\usepackage{booktabs}
\usepackage{graphicx}
\usepackage{tcolorbox}
\usepackage{xcolor}
\usepackage{colortbl}
\usepackage{tabularx}
\usepackage[T1]{fontenc}

\usepackage[utf8]{inputenc}

\usepackage{microtype}

\usepackage{inconsolata}

\usepackage{graphicx}

\usepackage[table]{xcolor}
\usepackage{amsmath}

\title{\textsc{KoVRE}: Training an Efficient Embedding Model for \\Korean Visual Document Retrieval}

\author{
Yongbin Choi$^{1}$, Gyuho Shim$^{2}$, Youngjoon Jang$^{2\dagger}$
\\
$^{1}$Kyung Hee University \quad $^{2}$Korea University \\
\texttt{yongbinchoi@khu.ac.kr} \\ 
\texttt{\{gjshim, dew1701\}@korea.ac.kr}
}

\newcommand\blfootnote[1]{%
  \begingroup
  \renewcommand\thefootnote{}\footnote{#1}%
  \addtocounter{footnote}{-1}%
  \endgroup
}

\begin{document}
\maketitle
\begin{abstract}
Visual Document Retrieval (VDR) directly matches text queries against document images, preserving visual and structural information that may be lost during text extraction. However, existing VDR models and training resources remain predominantly English-centric, while many high-performing systems rely on massive backbones or storage-intensive multi-vector representations. To address these limitations, we introduce \textbf{\textsc{KoVRE}}: \textbf{\textsc{Ko}}rean \textbf{\textsc{V}}isual Document \textbf{\textsc{R}}etrieval \textbf{\textsc{E}}mbedding, a single-vector retriever for Korean visual documents, alongside a comprehensive training recipe. We train the model on 708,729 Korean and English query--page pairs using positive-aware hard-negative mining and conduct controlled analyses of training-data composition, hard-negative treatment, and reranker-based knowledge distillation. Across Korean visual document retrieval benchmarks, our 2B model substantially improves over the base backbone model, outperforming both its 8B single-vector counterpart and a strong multi-vector baseline. These results demonstrate that targeted bilingual supervision and our carefully designed training strategies can produce a highly effective Korean VDR model across diverse document domains, without requiring a scaled-up backbone or multi-vector representations.
\end{abstract}

\section{Introduction}

\blfootnote{$^\dagger$Corresponding author.}

Retrieval-Augmented Generation (RAG) allows language models to ground their outputs in external documents, but its effectiveness depends on retrieving the right evidence~\citep{lewis2020retrieval, yu2025visrag, wang-etal-2025-vidorag, tanaka2025vdocrag}. Most document retrieval systems index text extracted from document pages through pipelines involving Optical Character Recognition (OCR) and layout detection~\citep{cho2024m3docrag, dong-etal-2025-mmdocir}. These pipelines are prone to errors and can discard visual cues encoded in page layouts, tables, figures, infographics, and spatial relationships. As a result, relevant information may become difficult to retrieve when its meaning depends on both textual and visual content~\citep{faysse2024colpali, ma-etal-2024-unifying}.

Visual document retrieval (VDR) addresses this limitation by directly matching text queries with rendered document pages~\citep{mace2025vidore, loison-etal-2026-vidore, wasserman-etal-2025-real, peng2025unidoc, shorten2026irpapers}. Because it operates on page images, VDR retains access to textual, visual, and structural evidence without relying solely on extracted text. Recent VDR models have achieved strong retrieval performance, but many rely on large vision-language backbones or multi-vector representations that store multiple token- or patch-level embeddings for each page~\citep{huang2025beyond, nomicembedmultimodal2025, xu2025llama, xiao2025metaembed, gunther-etal-2025-jina, moreira2026nemotron, li2026qwen3}. Although multi-vector representations enable fine-grained matching, they also increase storage and retrieval costs as the collection grows~\citep{ma-etal-2025-towards-storage}. A compact single-vector model therefore offers an attractive alternative when both retrieval quality and scalability matter.

However, progress in VDR has been largely driven by English-centric training data and evaluation benchmarks~\citep{osmulski2025miracl}. Models with general multilingual capabilities do not necessarily provide strong retrieval performance in Korean, and only a limited number of models perform competitively across Korean VDR benchmarks~\citep{lee2025sds, choi-etal-2026-kovidore}. Consequently, there is still a need for a compact visual document retriever that is explicitly adapted to Korean. Furthermore, how such a model should be trained remains underexplored, particularly with respect to training-language composition and how hard negatives are mined and treated, and how supervision from a teacher reranker can be effectively distilled into a compact retriever.



To address this gap, we adapt Qwen3-VL-Embedding-2B~\citep{li2026qwen3} for Korean visual document retrieval. We train the model on Korean and English query--page pairs with mined hard negatives and a Matryoshka multiple-negative ranking objective~\citep{kusupati2022matryoshka}. Alongside the model, we systematically examine the effects of 
training data composition, self-guide filtering, hardness weighting, score normalization in knowledge distillation, and embedding dimensionality.
Across Korean VDR benchmarks, our model substantially improves upon the original 2B model and outperforms both its 8B single-vector counterpart and a strong multi-vector model. These results demonstrate that well-designed training can produce a competitive Korean retriever without requiring a larger backbone or a multi-vector index. Our contributions are as follows:
\begin{itemize}
    \item We develop a compact single-vector retriever for Korean visual documents through 
    two-stage
    training, without modifying the underlying vision-language architecture.
    \item We provide a systematic analysis of the training components: language composition, addition of text dataset, hard-negative construction, self-guide filtering, hardness weighting, min-max scaling, and embedding dimensionality.
\end{itemize}

\section{Related Work}

\subsection{Visual Document Retrieval and Benchmarks}

Visual document retrieval represents document pages directly from their rendered images and retrieves them using textual queries. ColPali~\citep{faysse2024colpali} established a prominent late-interaction approach by adapting a vision--language model to produce multiple embeddings per page. Subsequent models such as ColNomic~\citep{nomicembedmultimodal2025} and Nemotron ColEmbed~\citep{xu2025llama, moreira2026nemotron} retain this multi-vector formulation to preserve token-level matching, but storing and scoring many vectors per page incurs substantial index storage and retrieval overhead~\citep{ma-etal-2025-towards-storage}. ColModernVBERT~\citep{teiletche2025modernvbert} addresses this cost from a different angle, designing a compact bidirectional encoder while retaining late interaction. A complementary direction compresses each query and page into a single dense vector, as seen in Nomic Embed Multimodal, released alongside ColNomic~\citep{nomicembedmultimodal2025}, and Qwen3-VL-Embedding~\citep{li2026qwen3}, while Jina Embeddings v4~\citep{gunther-etal-2025-jina} supports both formulations. Qwen3-VL-Embedding further supports variable output dimensions via Matryoshka Representation Learning (MRL)~\citep{kusupati2022matryoshka} and is released at multiple scales, making it a suitable foundation for studying compact retrieval.

Although early VDR resources focused primarily on English or European languages~\citep{mace2025vidore, loison-etal-2026-vidore}, recent benchmarks have introduced evaluation suites for Korean visual documents. SDS KoPub VDR~\citep{lee2025sds} evaluates structured public documents across textual, visual, and cross-modal queries, while KoViDoRe~\citep{choi-etal-2026-kovidore} provides multi-domain documents and multi-page targets alongside the Ko-VDR Train Public corpus. Together, these resources establish evaluation infrastructure for Korean VDR, while training data for the language remains limited in both scale and document diversity relative to its English counterparts.

\subsection{Training Techniques for Dense Retrievers}

The training techniques examined in this work were largely developed for text retrieval. Hard-negative mining strengthens contrastive training, and ANCE~\citep{xiong2020approximatenearestneighbornegative} improves negative hardness by sampling from a periodically refreshed global index. Harder negatives, however, raise the risk of false negatives, prompting strategies to detect and remove them: RocketQA~\citep{qu2021rocketqaoptimizedtrainingapproach} filters mined candidates using a cross-encoder, NV-Retriever~\citep{moreira2024nv} introduces positive-aware thresholds that discard candidates scoring too close to the positive, and GISTEmbed~\citep{solatorio2024gistembedguidedinsampleselection} masks in-batch candidates that a separate guide model scores above the positive. Rather than removing such negatives, LLaVE~\citep{lan2025llave} and EmbeddingGemma~\citep{vera2025embeddinggemma} reweight mined negatives by difficulty.

Distilling cross-encoder rerankers into bi-encoder retrievers is likewise well established. margin-MSE~\citep{hofstatter2020improving} matches score margins between teacher and student, while RocketQAv2~\citep{ren2023rocketqav2jointtrainingmethod} and ColBERTv2~\citep{santhanam2022colbertv2effectiveefficientretrieval} adopt listwise objectives that align normalized score distributions. Because reranker scores lie on an unbounded, query-dependent scale, these methods place teacher and student scores on a common scale before comparison. These techniques have been developed and validated mainly on text retrieval, and their behavior in visual document retrieval remains largely unexamined.

\input{tables/1_training_data}

\section{Training Method}

We train the retriever in two stages. 
In Stage~1, we perform Contrastive Learning (CL) on Korean and English visual document retrieval data, where the English data is included deliberately to prevent catastrophic forgetting of the backbone's pre-existing retrieval ability during Korean adaptation. We then perform reranker-based Knowledge Distillation (KD) using only the Korean data to further focus the model on Korean retrieval in Stage~2.

\subsection{Base Model and Representation}

We initialize our retriever from Qwen3-VL-Embedding-2B and largely follow its embedding training recipe~\citep{li2026qwen3}. The model accepts a text query or a rendered document page and maps the input to a shared dense space. We preserve the original architecture and use a single 2,048-dimensional vector to represent each page at full dimensionality. The query instruction is \texttt{"Find a document image that matches the given query."}, while document pages use the default instruction, \texttt{"Represent the user's input"}. Cosine similarity is used for both training and retrieval.

Following the original recipe, we preserve the aspect ratio of each page image and cap its resolution at 1,280 visual tokens, corresponding to approximately $1.3\times10^{6}$ pixels. Qwen3-VL-Embedding-2B supports user-defined output dimensions from 64 to 2,048. During training, we apply Matryoshka Representation Learning (MRL) at 2,048, 1,024, 768, 512, 256, and 128 dimensions~\citep{kusupati2022matryoshka}.

\subsection{Training Data Construction}
\paragraph{Stage~1 Training Data}
Table~\ref{tab:training_data} summarizes the datasets used in Stage~1, listing the original dataset names and corresponding numbers of Q--D pairs. The Korean component combines the public training resource released with KoViDoRe~\citep{choi-etal-2026-kovidore} and an additional private Korean collection from AI Hub.\footnote{\url{https://aihub.or.kr/}} The English component combines five existing VDR resources covering reports, slides, tables, and other visually structured pages~\citep{faysse2024colpali, yu2025visrag, wasserman-etal-2025-real, zhu-etal-2021-tat, llamaindex2025vdrmultilingual}. After mining and filtering, the Korean and English components retain 406,945 and 301,784 Q--D pairs, respectively. Each retained pair consists of one text query, one positive page image, and seven mined negative page images.

We mine seven hard negatives per query using Qwen3-VL-Embedding-8B. Within each source dataset, we rank pages by cosine similarity after excluding known positives. Following NV-Retriever~\citep{moreira2024nv}, we also exclude candidates whose similarity exceeds 95\% of the annotated positive similarity to reduce false negatives. We retain the seven highest-ranked eligible pages and discard pairs with fewer than seven candidates. We additionally remove weakly aligned query--positive pairs. Based on separate inspection of the Korean and English score distributions and manually reviewed samples, we retain a query only if at least one positive scores above 0.3. This filtering removes 9,359 Korean and 21,659 English pairs, yielding the totals reported in Table~\ref{tab:training_data}.

\paragraph{Stage~2 Training Data} 
Stage~2 uses only the Korean component of this corpus, keeping the same queries and positive pages but replacing binary hard negatives with soft distillation targets from a reranker teacher. 
For each query, we assemble a candidate pool of 64 pages: the 32 highest-ranked embedding-hard negatives mined with Qwen3-VL-Embedding-8B and 32 pages randomly sampled from the corpus. We then score every query--page pair with Qwen3-VL-Reranker-8B~\citep{li2026qwen3}. These reranker scores act as the teacher relevance labels that the student is distilled toward. 
From the 64 scored candidates we retain the eight negatives with the highest teacher scores, concentrating supervision on the pages the teacher finds most confusable with the positive. Each Stage~2 instance thus consists of one query and a candidate set containing one positive and eight negative pages, together with a teacher-score vector over the nine candidates. This yields a total of 265,311 query-level distillation instances.

\subsection{Stage 1: Contrastive Learning}

In Stage 1, we optimize an InfoNCE objective over paired positives, explicitly mined hard negatives, and in-batch negatives. Let $q_i$ denote a query, $d_i^{+}$ its paired positive page, $\mathcal{H}_i$ the set of seven mined hard negatives, and $\mathcal{B}_i$ the set of in-batch negatives. We define the full negative set as $\mathcal{D}_i^{-}=\mathcal{H}_i\cup\mathcal{B}_i$. Before self-guide filtering and hardness weighting, the loss is
\begin{equation*}
\mathcal{L}_{i}^{\mathrm{CL}}=-\log
\frac{\exp(s(q_i,d_i^{+})/\tau)}
{\sum_{d\in\{d_i^{+}\}\cup\mathcal{D}_i^{-}}
\exp(s(q_i,d)/\tau)},
\end{equation*}
where $s(\cdot,\cdot)$ is cosine similarity and $\tau$ is a temperature parameter.

We apply self-guide filtering to all negatives in $\mathcal{D}_i^{-}$. Our self-guide setting of $-0.1$ excludes a negative $d\in\mathcal{D}_i^{-}$ from the denominator when
\begin{equation*}
\mathrm{sg}\!\left(s(q_i,d)\right)>
\mathrm{sg}\!\left(s(q_i,d_i^{+})\right)+0.1,
\end{equation*}
where $\mathrm{sg}$ denotes stop-gradient. The paired positive is never masked. This rule reduces the influence of potential false negatives that the model scores substantially higher than the annotated positive.

We additionally adopt the hardness-weighted contrastive learning strategy used in LLaVE~\citep{lan2025llave} and EmbeddingGemma~\citep{vera2025embeddinggemma}. For each explicitly mined hard negative $h\in\mathcal{H}_i$, we add a stop-gradient hardness term to its contrastive logit:
\begin{equation*}
\ell(q_i,h)=\frac{s(q_i,h)}{\tau}
+\alpha\,\mathrm{sg}\!\left(s(q_i,h)\right),
\end{equation*}
where $\alpha$ controls the hardness-weighting strength. We set $\alpha=2$ and apply the additional term only to the seven mined hard negatives in $\mathcal{H}_i$; in-batch negatives in $\mathcal{B}_i$ retain the unweighted logit $s(q_i,d)/\tau$. After self-guide filtering, this weighted logit replaces the standard InfoNCE logit for each remaining hard negative. 

\input{tables/2_main}

We apply the resulting contrastive objective jointly across the Matryoshka dimensions. Let $\mathcal{M}=\{m_k\}_{k=1}^{K}$ denote the set of training dimensions specified above. For an input $x$, its representation at dimension $m\in\mathcal{M}$ is obtained by truncating and normalizing the full embedding:
\begin{equation*}
\mathbf{z}_{x}^{(m)}
=
\frac{\mathbf{z}_{x,1:m}}
{\lVert\mathbf{z}_{x,1:m}\rVert_2},
\qquad
s_m(q,d)
=
{\mathbf{z}_{q}^{(m)}}^\top\mathbf{z}_{d}^{(m)}.
\end{equation*}
For each $m\in\mathcal{M}$, we compute the contrastive loss described above using $s_m(\cdot,\cdot)$ with the same self-guide filtering and hardness weighting. The final Stage~1 objective is
\begin{equation*}
\mathcal{L}^{\mathrm{Stage1}}
=
\frac{1}{N}
\sum_{i=1}^{N}
\sum_{m\in\mathcal{M}}
\mathcal{L}_{i}^{\mathrm{CL},(m)},
\end{equation*}
where $\mathcal{L}_{i}^{\mathrm{CL},(m)}$ denotes the contrastive loss for query $i$ at dimension $m$, and $N$ is the number of queries in the contrastive batch.

\subsection{Stage 2: Knowledge Distillation}

In Stage 2, we optimize the KL-Divergence objective over the teacher scores. For each query $q_i$, let $\mathcal{C}_i$ denote the candidate set containing its paired positive and eight mined pages. We use $t_{ij}$ and $u_{ij}$ to denote the teacher reranker score and student cosine similarity for candidate $d_j\in\mathcal{C}_i$, respectively.

Because the teacher and student scores have different numerical ranges, we independently apply min--max normalization to each score vector within the candidate set:

\begin{equation*}
\resizebox{1.0\columnwidth}{!}{$\displaystyle
\bar{x}_{ij}=
\frac{x_{ij}-\min_{k\in\mathcal{C}_i}x_{ik}}
{\max_{k\in\mathcal{C}_i}x_{ik}-\min_{k\in\mathcal{C}_i}x_{ik}+\epsilon}, \quad x\in\{t,u\},
$}
\end{equation*}

where $\epsilon$ is a small constant for numerical stability. We convert the normalized scores into distributions over the candidate set:
\begin{equation*}
p_{ij}^{T}=
\frac{\exp(\bar{t}_{ij})}
{\sum_{k\in\mathcal{C}_i}\exp(\bar{t}_{ik})},
\quad
p_{ij}^{S}=
\frac{\exp(\bar{u}_{ij})}
{\sum_{k\in\mathcal{C}_i}\exp(\bar{u}_{ik})}.
\end{equation*}
The student minimizes the KL divergence from the teacher distribution to the student distribution:
\begin{equation*}
\mathcal{L}_{i}^{\mathrm{KD}}
=\mathrm{D}_{\mathrm{KL}}\!\left(p_i^{T}\,\|\,p_i^{S}\right)
=\sum_{j\in\mathcal{C}_i}p_{ij}^{T}
\log\frac{p_{ij}^{T}}{p_{ij}^{S}}.
\end{equation*}

As in Stage~1, we apply this objective across the Matryoshka dimensions $\mathcal{M}$, deriving the student distribution from $s_m(\cdot,\cdot)$ at each dimension $m$ while the teacher distribution is shared across dimensions. The final Stage~2 objective is
\begin{equation*}
\mathcal{L}^{\mathrm{Stage2}}
=
\frac{1}{N}
\sum_{i=1}^{N}
\sum_{m\in\mathcal{M}}
\mathcal{L}_{i}^{\mathrm{KD},(m)},
\end{equation*}
where $\mathcal{L}_{i}^{\mathrm{KD},(m)}$ denotes the distillation loss for query $i$ at dimension $m$.

\subsection{Implementation Details}

In Stage~1, we train the full 2B model for one epoch using bfloat16 mixed precision on two B200 GPUs. The per-device batch size is 128 with two gradient-accumulation steps. We apply GradCache~\citep{gao2021scaling} with a mini-batch size of 8, decoupling embedding computation from loss computation to support a larger effective contrastive batch under the available GPU memory. We use a learning rate of $2\times10^{-5}$, cosine decay, and a warmup proportion of 0.1. Duplicate examples are excluded within a batch, and datasets are sampled proportionally. Korean and English examples remain in separate datasets and are not mixed within the same batch. In Stage~2, we train the student model on eight RTX PRO 6000 GPUs. The per-device batch size is 8 with two gradient-accumulation steps, yielding an effective batch size of 128. We use a learning rate of $1\times10^{-6}$ in this stage.

\section{Experimental Setup}

\subsection{Benchmarks and Metrics}

We evaluate Korean visual document retrieval in two complementary settings. KoViDoRe contains Korean queries over public and enterprise-style documents from four domains: cybersecurity, economics, energy, and human resources. A query may correspond to multiple relevant pages, allowing the benchmark to measure ranking quality across diverse document domains. SDS KoPub VDR contains Korean public documents with textual, visual, and cross-modal query categories. This setting evaluates out-of-distribution transfer in Korean visual document retrieval. We report nDCG@10 as the main evaluation metric, following the Massive Multilingual Text Embedding Benchmark (MMTEB)~\citep{enevoldsen2025mmteb}.

\subsection{Baselines}

We select Qwen3-VL-Embedding-8B and jina-embeddings-v4 as baselines because both show strong performance on KoViDoRe and SDS KoPub VDR. We additionally include Qwen3-VL-Embedding-2B, the original backbone from which our model is initialized, to measure the effect of Korean adaptation. For jina-embeddings-v4, we evaluate both its single-vector and multi-vector representations. We also include jina-clip-v2~\citep{koukounas2024jina}, jina-v5-omni-nano, jina-v5-omni-small~\citep{honicke2026jina}. All retrieval evaluations are conducted using the MTEB framework~\citep{muennighoff-etal-2023-mteb}.~\footnote{\url{https://github.com/embeddings-benchmark/mteb}}

\section{Results and Analysis}
\label{sec:results}

We first report the performance of \textsc{KoVRE} after Stage~1 and Stage~2 relative to single-vector and multi-vector baselines. We then examine the main components of the training recipe through controlled ablations on training-data composition, negative treatment, score normalization in knowledge distillation, and embedding dimensionality under Matryoshka representation learning (MRL). Unless otherwise stated, each experiment varies only the component under study.

\subsection{Main Results}

Table~\ref{tab:main} summarizes the overall results across the evaluated benchmarks. Stage~1 improves consistently over the original Qwen3-VL-Embedding-2B model and already approaches the strongest baseline overall, showing that bilingual contrastive training and hard-negative supervision provide an effective adaptation signal for Korean VDR. Stage~2 further improves performance on every benchmark, indicating that reranker-based knowledge distillation transfers additional ranking information beyond the contrastive objective.

The final \textsc{KoVRE} model achieves the best aggregate performance and the highest KoViDoRe average, with particular strengths in the economic and human-resources domains. Qwen3-VL-Embedding-8B remains strongest on cybersecurity, while multi-vector jina-embeddings-v4 leads on energy and SDS KoPub VDR. These domain-level differences suggest that no model dominates every document distribution, but \textsc{KoVRE} provides the most balanced performance across the evaluated Korean benchmarks. Its overall advantage over both a larger single-vector model and a strong multi-vector model further indicates that targeted training can be more consequential than increasing model size or adopting a storage-intensive representation.

\begin{figure}[t!]
\centering
\includegraphics[width=1.0\linewidth]{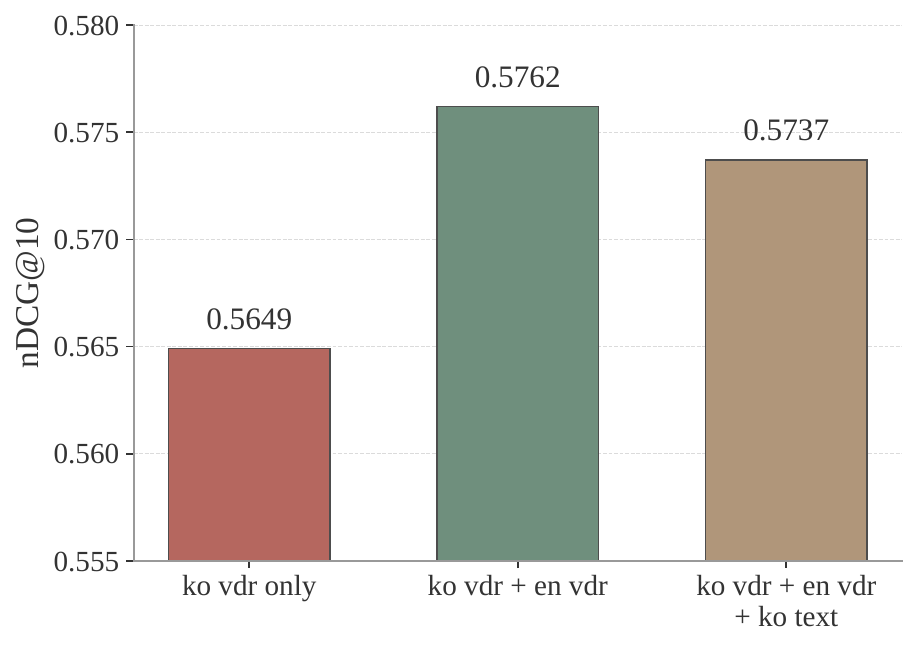}
\caption{Effect of training-data composition on Korean VDR performance.}
\label{fig:training_data_composition}
\end{figure}

\subsection{Effect of Training Data Composition}

We compare three training mixtures: Korean VDR data only, Korean and English VDR data, and the bilingual VDR mixture augmented with an additional Korean text dataset. As illustrated in Figure~\ref{fig:training_data_composition}, the bilingual VDR mixture performs better overall than the Korean-only configuration. Mixing English examples may provide a rehearsal-like signal that helps preserve the backbone's pre-existing multilingual retrieval capability during Korean adaptation~\citep{mhamdi-may-2024-leitner, huang-etal-2024-mitigating}. 

To evaluate the effect of an additional text dataset, we use the Korean retrieval dataset \texttt{nlpai-lab/ko-triplet-v1.0}~\citep{jang2024koe5} and select seven hard negatives per query using Qwen3-Embedding-8B. These examples are added to the bilingual VDR corpus but organized into separate batches from the visual document examples. Adding the Korean text dataset reduces performance on both benchmarks and the overall macro average relative to the bilingual VDR mixture. Additional text supervision therefore provides no benefit under the current batching and sampling strategy. We consequently retain only the Korean and English VDR datasets in the final Stage~1 training mixture.

\subsection{Negative Treatment}

\input{tables/3_hardness_selfguide}

We jointly examine self-guide filtering and the strength of hardness weighting applied to the seven explicitly mined hard negatives. Table~\ref{tab:hardness_selfguide} compares a configuration without self-guide filtering with two configurations that use a self-guide threshold of $-0.1$. At a fixed hardness weight~($\alpha=2.0$), adding self-guide filtering improves SDS KoPub VDR and the overall result while preserving KoViDoRe performance. By contrast, increasing the hardness weight yields only a negligible gain on KoViDoRe and weakens performance on SDS KoPub VDR, suggesting that excessive emphasis on mined hard negatives can reduce transfer to an unseen document distribution. Taken together, these results support using $\alpha=2.0$ with a self-guide threshold of $-0.1$ in the final Stage~1 configuration.

\begin{figure}[t!]
\centering
\includegraphics[width=1.0\linewidth]{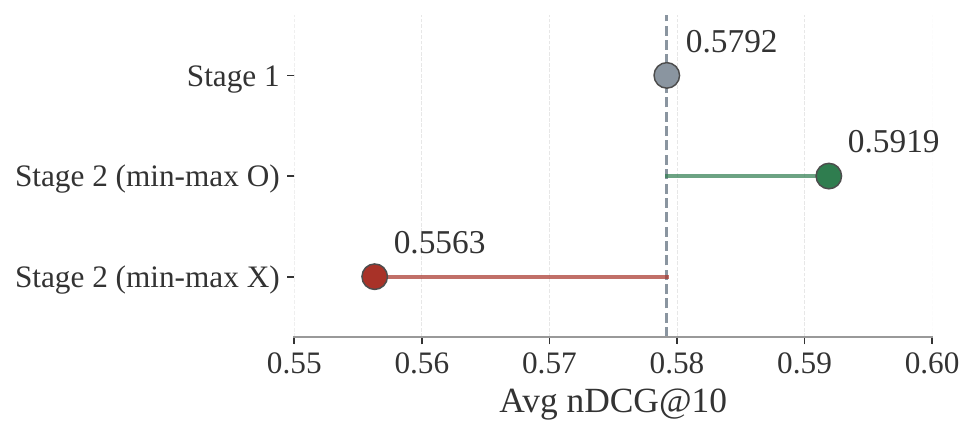}
\caption{Effect of per-row min--max score normalization on Stage~2 distillation, measured by average nDCG@10.}
\label{fig:minmax_ablation}
\end{figure}

\subsection{Score Normalization for Knowledge Distillation}

During Stage~2, we distill the reranker teacher into \textsc{KoVRE} by matching, for each query, the student's distribution over the positive and eight negatives to that of the teacher. Because the reranker assigns scores on an unbounded, query-dependent scale, applying a softmax directly to the raw teacher scores yields sharply peaked targets that collapse onto the single highest-scoring page, discarding the graded relevance information that motivates distillation. To address this, we apply per-row min--max normalization that rescales each query's student and teacher scores to $[0,1]$ before applying the softmax, placing the two distributions on a common scale while preserving the shape of the teacher's relevance ordering.

Figure~\ref{fig:minmax_ablation} shows that per-query min--max normalization is critical under our tested distillation configuration. Without min--max normalization, Stage~2 distillation is actively harmful: the overall score drops to 0.5563, well below the 0.5792 of the Stage~1 checkpoint. With min--max normalization, distillation instead improves the model to 0.5919. Score normalization is therefore a prerequisite for effective knowledge distillation in our setting.

\subsection{Performance of Matryoshka Dimensions}
\begin{figure}[t!]
\centering
\includegraphics[width=1.0\linewidth]{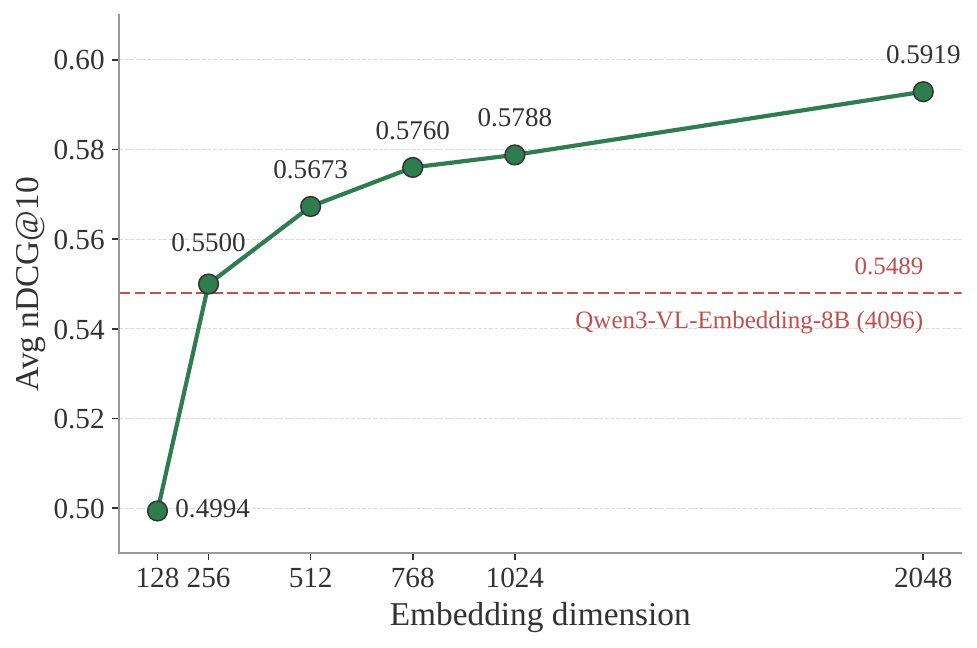}
\caption{Overall nDCG@10 of \textsc{KoVRE} across MRL dimensions. The dashed line denotes Qwen3-VL-Embedding-8B with a 4,096-dimensional representation.}
\label{fig:mrl_dim_ablation}
\end{figure}

Figure~\ref{fig:mrl_dim_ablation} shows the overall retrieval performance of \textsc{KoVRE} across the embedding dimensions used for Matryoshka representation learning. \textsc{KoVRE} retains strong retrieval performance even as the embedding dimension is substantially reduced, indicating that MRL effectively preserves task-relevant information in compact representations. Notably, the 256-dimensional \textsc{KoVRE} representation slightly outperforms Qwen3-VL-Embedding-8B with 4,096 dimensions while using only one-sixteenth as many embedding dimensions. Because the two models also differ in size and training data, this comparison should be interpreted as evidence for the practical compactness of the learned representation rather than as an isolated effect of dimensionality.

\section{Conclusion}

We introduce \textsc{KoVRE}, a compact single-vector retriever for Korean visual documents adapted from Qwen3-VL-Embedding-2B. \textsc{KoVRE} is trained in two stages: contrastive learning on 708,729 Korean and English query--page pairs with positive-aware hard-negative mining and targeted loss design, followed by reranker-based knowledge distillation on the Korean subset, for which per-query min--max score normalization proves essential. Across Korean VDR benchmarks, the adapted 2B model substantially improves over the original checkpoint, outperforms the corresponding 8B single-vector model, and slightly exceeds a strong 4B multi-vector baseline in aggregate performance, despite using a smaller backbone and a single-vector representation that remains competitive even at a fraction of its full dimensionality. We hope that our training recipe and the accompanying analyses serve as a practical starting point for building visual document retrievers in other languages that remain underserved by current models.

\section*{Limitations}

The Korean training corpus contains substantially fewer distinct page images than its English counterpart, despite providing more Q--D pairs (35,815 versus 226,714 pages). Korean supervision is therefore concentrated on a relatively limited pool of pages, restricting the diversity of layouts, document types, and visual structures observed during training. This limited exposure may constrain generalization to Korean documents with page formats that are underrepresented in the training corpus. Future work should expand the diversity of Korean page images rather than increasing only the number of queries associated with existing pages.

\bibliography{custom}

\clearpage

\appendix



\end{document}

%% file: tables/1_training_data.tex
\begin{table}[t]
\centering
\small
\resizebox{\columnwidth}{!}{%
\begin{tabular}{lrr}
\hline
Dataset & Pages & Q--D Pairs \\
\hline
\rowcolor{gray!15}
\multicolumn{3}{c}{\textit{Korean}} \\
\texttt{ko-vdr-train-public} & 7,548 & 310,226 \\
\texttt{ko-vdr-train-private} & 28,267 & 118,624 \\
\textbf{Subtotal} & \textbf{35,815} & \textbf{428,850} \\
\textit{After hard-negative selection} & & 416,304 \\
\textit{After positive refinement} & & \textbf{406,945} \\
\hline
\rowcolor{gray!15}
\multicolumn{3}{c}{\textit{English}} \\
\texttt{VisRAG-Ret-Train-In-domain-data} & 84,417 & 122,752 \\
\texttt{vdr-multilingual-train} (en) & 53,335 & 53,512 \\
\texttt{REAL-MM-RAG\_FinTabTrainSet\_rephrased} & 48,206 & 48,206 \\
\texttt{colpali\_train\_set} & 38,866 & 118,195 \\
\texttt{tatdqa\_train} & 1,890 & 13,251 \\
\textbf{Subtotal} & \textbf{226,714} & \textbf{355,916} \\
\textit{After hard-negative selection} & & 323,443 \\
\textit{After positive refinement} & & \textbf{301,784} \\
\hline
\textbf{Total} & \textbf{262,529} & \textbf{708,729} \\
\hline
\end{tabular}%
}
\caption{Composition of the bilingual corpus used for contrastive learning in Stage~1. Pages denote deduplicated page images in the hard-negative corpora. Source subtotals give the number of query--positive pairs before filtering. 
}
\label{tab:training_data}
\end{table}

%% file: tables/2_main.tex
\begin{table*}[t]
\centering
\resizebox{\textwidth}{!}{%
\renewcommand{\arraystretch}{1.1}
\begin{tabular}{lc|cccccc||c}
\toprule
\multirow{2}{*}{Model} & \multirow{2}{*}{\# params} & \multicolumn{5}{c}{KoViDoRe} & \multirow{2}{*}{SDSKoPub} & \multirow{2}{*}{OVR} \\ \cmidrule(lr){3-7}
& & Cybersecurity & Economic & Energy & HR & AVG & & \\ \midrule
jina-clip-v2 & 0.9B & 0.1993 & 0.0011 & 0.1096 & 0.0294 & 0.0849 & 0.0732 & 0.0825 \\ 
jina-v5-omni-nano & 1B & 0.4404 & 0.0640 & 0.2017 & 0.0695 & 0.1939 & 0.0961 & 0.1743 \\ 
jina-v5-omni-small & 2B & 0.4357 & 0.0744 & 0.2380 & 0.1029 & 0.2128 & 0.1902 & 0.2082 \\ 
Qwen3-VL-Embedding-2B & 2B & 0.6111 & 0.1592 & 0.4123 & 0.1842 & 0.3417 & 0.4285 & 0.3591 \\ 
Qwen3-VL-Embedding-8B & 8B & \textbf{0.7809} & 0.2373 & 0.6360 & 0.3613 & 0.5039 & 0.7293 & 0.5489 \\ 
jina-embeddings-v4 (single-vector) & 4B & 0.7280 & 0.2058 & 0.6273 & 0.4106 & 0.4929 & 0.7222 & 0.5388 \\ 
jina-embeddings-v4 (multi-vector) & 4B & \underline{0.7714} & 0.2359 & \textbf{0.6752} & 0.4799 & 0.5406 & \textbf{0.7802} & \underline{0.5885} \\ \midrule
\textbf{\textsc{KoVRE} (Stage 1)} & 2B & 0.7444 & \underline{0.2797} & 0.6506 & \underline{0.5002} & \underline{0.5437} & 0.7214 & 0.5792 \\
\textbf{\textsc{KoVRE} (Stage 1 + Stage 2)} & 2B & 0.7627 & \textbf{0.2987} & \underline{0.6576} & \textbf{0.5082} & \textbf{0.5568} & \underline{0.7324} & \textbf{0.5919} \\ \bottomrule
\end{tabular}%
}
\caption{nDCG@10 results on Korean VDR benchmarks. AVG is averaged over the four KoViDoRe domains, and OVR is the macro-average over these four domains and SDS KoPub VDR.}
\label{tab:main}
\end{table*}

%% file: tables/3_hardness_selfguide.tex
\begin{table}[t]
\centering
\small
\resizebox{\columnwidth}{!}{%
\renewcommand{\arraystretch}{1.1}
\begin{tabular}{cc|cc||c}
\hline
\textbf{($\alpha$)} & \textbf{self-guide} & \textbf{KoViDoRe} & \textbf{SDS KoPub VDR} & \textbf{OVR} \\
\hline
2.0 & X & 0.5426 & 0.7100 & 0.5761 \\
2.0 & -0.1 & 0.5437 & 0.7214 & \textbf{0.5792} \\
5.0 & -0.1 & 0.5439 & 0.7094 & 0.5770 \\
\hline
\end{tabular}%
}
\caption{Ablation study on hardness weight ($\alpha$) and self-guide.}
\label{tab:hardness_selfguide}
\end{table}